# Handwritten Arabic Numeral Recognition using a Multi Layer Perceptron

Nibaran Das[+], Ayatullah Faruk Mollah[*], Sudip Saha[*], Syed Sahidul Haque[*]

[+] Computer Science and Engineering Department,
Techno India, Kolkata-700091, India
[*] Computer Science and Engineering Department,
Jadavpur University, Kolkata-700032, India

[+]Corresponding Author's email: nibs_breath@yahoo.com

**Abstract**

*Handwritten numeral recognition is in general a benchmark problem of Pattern Recognition and Artificial Intelligence. Compared to the problem of printed numeral recognition, the problem of handwritten numeral recognition is compounded due to variations in shapes and sizes of handwritten characters. Considering all these, the problem of handwritten numeral recognition is addressed under the present work in respect to handwritten Arabic numerals. Arabic is spoken throughout the Arab World and the fifth most popular language in the world slightly before Portuguese and Bengali. For the present work, we have developed a feature set of 88 features is designed to represent samples of handwritten Arabic numerals for this work. It includes 72 shadow and 16 octant features. A Multi Layer Perceptron (MLP) based classifier is used here for recognition handwritten Arabic digits represented with the said feature set. On experimentation with a database of 3000 samples, the technique yields an average recognition rate of 94.93% evaluated after three-fold cross validation of results. It is useful for applications related to OCR of handwritten Arabic Digit and can also be extended to include OCR of handwritten characters of Arabic alphabet*

## 1. Introduction

The automatic recognition of digits on scanned images has wide commercial importance. It has applications in OCR systems, automatic pin code recognition, cheque reading, collecting data from filled in forms. Though there are some commercially available softwares, mainly for printed character recognition of some languages, But the success yet to be extended for hand written characters. Such technique is to facilitate smoother interaction between man and machine. The technique of Handwritten Arabic numeral recognition can contribute tremendously to the development of a complete OCR system. Therefore OCR of handwritten numerals is still an active area of research.

The past work on OCR of handwritten alphabets and numerals has mostly found to concentrate on Roman script[3] related to English and some other European languages, and scripts related to Asian languages like Chinese [2], Korean, Japanese. Among Indian scripts, Devnagari, Tamil, Oriya and Bangle have started to receive attention for OCR related research in recent years. Compared to these, Arabic is one of the major languages in the world. It is spoken in a large area including North Africa, most of the Arabian Peninsula and other parts of the Middle East. About 500 million peoples speak in this language. Rankwise it is the fifth most popular language in the world. Popularity wise Portugese and Bengali slightly trail behind Arabic. Arabic is the official language of around 24 countries like Algeria, Baharain, Egypt etc and also the national language of Mali, Senegal, Somali. More over Arabic is the liturgical language of Islam. It is sometimes difficult to translate Islamic concepts, and concepts specific to Arab culture, with out using original Arabic terminology. The Arabic script has been adapted to such diverse languages as Persian (Farsi), Turkish, Spanish, Urdu, and Swahili. In spite of that, OCR of handwritten Arabic script including numerals has not so far received sufficient attention. Majority of the past work related to OCR of Arabic script was done with the printed characters[1].

Arabic words and characters within the words are written from right to left. But Arabic number is written from left to right. In the light of above facts, the present work concentrates on the development of an MLP based pattern classifier for recognition of handwritten Arabic numerals with a feature set of 88 features. There is also enough scope for extension of this work to include handwritten characters of Arabic alphabet . Due to non-availability of some standardized data set, recognition performances of the present work cannot be compared with others. However, for currently available data set, the work shows satisfactory recognition performances with some specific suggestions to be tried in future for further performance enhancement.

|  | . | ١ | ٢ | ٣ | ٤ |
|---|---|---|---|---|---|
| Roman Digit | 0 | 1 | 2 | 3 | 4 |
|  | ٥ | ٦ | ٧ | ٨ | ٩ |
| Roman Digit | 5 | 6 | 7 | 8 | 9 |

Fig.1. The decimal digit set of Arabic script

## 2 The Feature Sets

Choice of suitable features for pattern classes, as mentioned before, is a domain specific design Issue. In the present work, a feature set of 88 features are designed for classification of handwritten Arabic digit patterns. The feature set consists of two types of features. The features constituting this set are so selected that their values remain close to each other for the patterns of the same class and differ appreciably for the patterns of different classes. For extraction of features from the digit images, each digit image is first enclosed within minimal bounding boxes and then scaled to 32X32 pixel sizes. The scaled images, which are defined with gray scale pixel values, are finally converted to binary images through thresholding.

### 2.1. Shadow Features

Shadow features are computed by considering the lengths of projections of a digit image, as shown in Fig. 2, on the four sides and eight octant dividing sides of the minimal bounding box enclosing the same. Considering the lengths of projections on three sides of each such octant, 24 shadow features are extracted from each window of the digit image, which is divided into eight octants inside the minimal box. Each value of the shadow feature so computed is to be normalized by dividing it with the maximum possible length of the projections on the respective side. Here we have finally considered 3 overlapping windows, each producing 24 shadow features. Thus 72 features are considered with 3 overlapping windows.

### 2.2. Centroid features

Coordinates of centroids of black pixels in all the 8 octants of a digit image are considered to add 16 features in all to the feature set. Fig. 3(a-b) shows approximate locations of all such centroids on two different digit images. It is noteworthy how these features can be of help to distinguish the two images.

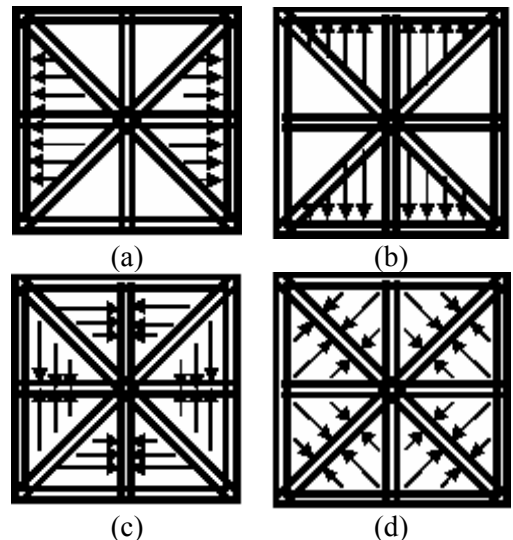

Fig 2. An illustration for shadow features.
(a-d) Direction of fictitious light rays as assume for taking the projection of an image segment on each side of all octants.

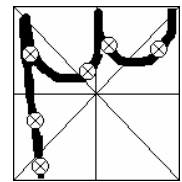

Fig. 3 An Illustration of the 16 Centroid features

## 3. The MLP Classifier

In the present work, an MLP classifier is employed for recognition of unknown digit patterns using the 88 element feature set. The MLP is a special kind of Artificial Neural Network (ANN). ANNs are developed to replicate *learning* and *generalization* abilities of human's behaviour with an attempt to model the functions of *biological neural networks* of the human brain.

Nowadays MLP is the mostly used classifier in the field of handwritten character recognition among the researcher [4].Architecturally, an MLP is a feed-forward layered network of *artificial neurons*. Each artificial neuron in the MLP computes a *sigmoid function* of the weighted sum of all its inputs. An MLP consists of one *input layer*, one *output layer* and a number of *hidden* or intermediate *layers*, as shown in Fig 4. The output from every neuron in a layer of the MLP is connected to all inputs of each neuron in the immediate next layer of the same. Neurons in the input layer of the MLP are all basically dummy neurons as they are used simply to pass on the input to the next layer just by computing an identity function each.

The numbers of neurons in the input and the output layers of an MLP are chosen depending on the problem to be solved. The number of neurons in other layers and the number of layers in the MLP are all determined by a trial and error method at the time of its *training*. An ANN requires training to learn an unknown input-output relationship to solve a problem.

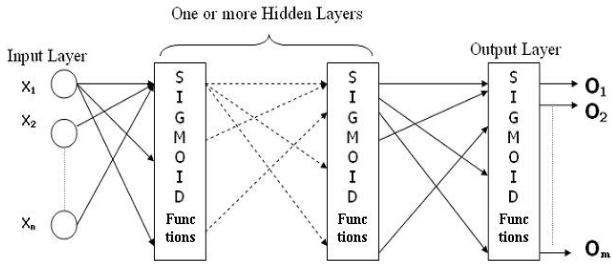

**Fig. 4.** A block diagram of an MLP shown as a feed forward neural network.

Depending on the models of ANNs, training is performed either under supervision of some teacher (i.e., with labeled data of known input-output responses) or without supervision. The MLP to be used for the present work requires supervised training. During training of an MLP *weights* or strengths of neuron-to-neuron connections, also called *synapses*, are iteratively tuned so that it can respond appropriately to all training data and also to other data, not considered at the time of training. Learning and generalization abilities of an ANN is determined on the basis of how best it can respond under these two respective situations.

The MLP classifier designed for the present work is trained with the Back Propagation (BP) algorithm. It minimizes the *sum of the squared errors* for the training samples by conducting a *gradient descent* search in the *weight space*. The number neurons in a hidden layer in the same are also adjusted during its training.

The problem of *pattern classification* involves two successive transformations as follows:

$$M \to F \to D$$

Where, M, F and D stand for the measurement space, the feature space and the decision space respectively. Once a feature set is fixed up, it is left with the design of a mapping ($\delta$) as follows:

$$\delta: F \to D$$

ANNs with their learning and generalization abilities can approximate a general class of functions given below.

$$f: \mathbb{R}^n \to \mathbb{R}$$

Pattern classification with ANNs requires approximating $\delta$ as a *discrete valued function* shown below.

$$\delta: \mathbb{R}^n \to \{1,2,..m\}$$

where, n and m denotes the number of features and the number of pattern classes respectively. So an ANN based pattern classifier requires n number of neurons in the input layer and m number of neurons in the output layer. Conventionally 1-out-of-m representation is used for its output.

## 4. Results and Discussion

For preparation of the *training* and the *test sets* of samples, a database of 3000 digit samples is formed by collecting optically scanned handwritten characters specimens of 10 symbols from each of 300 people of different age groups and sexes. A *training set* of 2000 samples and a *test set* of 1000 samples are then formed. All these samples are scaled to 32X32 pixel images first and then converted to *binary images* through thresholding

| | | | | | |
|---|---|---|---|---|---|
| | ش | ү | سر | ৪ | ع |
| True Class | ৩ | ৭ | ৩ | ৬ | ৪ |
| | (a) | (b) | (c) | (d) | (e) |

**Fig. 5 (a-e).** Some test samples classified truly by the MLP classifier.

| | | | | | |
|---|---|---|---|---|---|
| | ٧ | ۴ | μ | ٩ | ۲ |
| True Class | ৮ | ৬ | ২ | ৬ | ৭ |
| Recognized Class | ৬ | ৯ | ৩ | ৯ | ২ |
| | (a) | (b) | (c) | (d) | (e) |

**Fig. 6 (a-e).** Some test samples misclassified by the MLP classifier.

For the present work, a single layer MLP, i.e., an MLP with one hidden layer is chosen. This is mainly to keep the computational requirement of the MLP low without affecting its function approximation capability. According to Universal Approximation theorem [6] a single hidden layer is sufficient to compute a uniform approximation to a given training set.

To design an MLP for classification of handwritten alphabetic characters, several runs of BP algorithm with *learning rate* ($\eta$) = 0.8 and *momentum term* ($\alpha$)=0.7 are executed for different numbers of neurons in its hidden layer. Recognition performances of the MLP on the test sets observed from this

experimentation are given in Table1 with three fold cross validation of results

Curves showing variation of the Recognition performance of the MLP, for the three folds of test samples with increase in the number of neurons in its hidden layer are plotted in Fig. 7 from the Table 1. It is required to fix up the number of neurons in the hidden layer of MLP so that it can show the optimal recognition performance on the test set.

Recognition performances of the MLP, as observed from the curve shown in Fig. 7, initially rise as the number of neurons in the hidden layer is increased and fall after the same crosses some limiting value. It reflects the fact that for some fixed training and test sets, learning and generalization abilities of the MLP improve as the number of neurons in its hidden layer as increases up to certain limiting value and any further increase in the number of neurons in the hidden layer thereafter degrades the abilities. It is called the *over-fitting* problem.

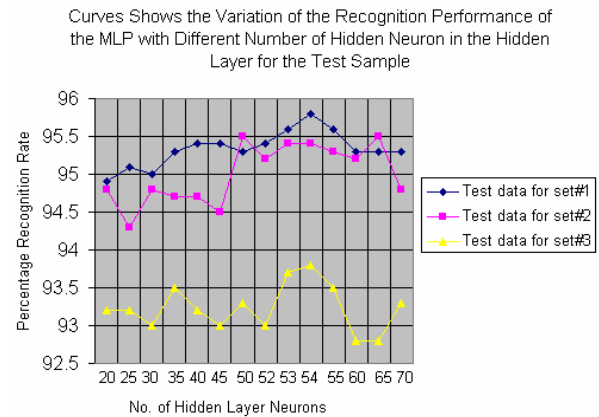

**Fig. 7.** Curves show variation of recognition performances of the MLP as the number of neurons in its hidden layer is increased

## Acknowledgement

Authors are thankful to the "Center for Microprocessor Application for Training Education and Research", "Project on Storage Retrieval and Understanding of Video for Multimedia" and Computer Science & Engineering Department, Jadavpur University, for providing infrastructural facilities during progress of the work. One of the authors, Mr. Nibaran Das, is thankful to Techno India for kindly permitting him to carry on the research work.

| Table 1. Recognition performances of the MLP with different numbers of neurons in the hidden layers ||||
|---|---|---|---|
| **No of Hidden neurons** | **Percentage recognition rate of the MLP on test samples** |||
| | *Fold#1* | *Fold#2* | *Fold#3* |
| 20 | 94.90 | 94.80 | 93.20 |
| 25 | 95.10 | 94.30 | 93.20 |
| 30 | 95.00 | 94.80 | 93.00 |
| 35 | 95.30 | 94.70 | 93.50 |
| 40 | 95.40 | 94.70 | 93.20 |
| 45 | 95.40 | 94.50 | 93.00 |
| 50 | 95.30 | **95.50** | 93.30 |
| 52 | 95.40 | 95.20 | 93.00 |
| 53 | 95.60 | 95.40 | 93.70 |
| 54 | **95.80** | 95.40 | **93.80** |
| 55 | 95.60 | 95.30 | 93.50 |
| 60 | 95.30 | 95.20 | 92.80 |
| 65 | 95.30 | 95.50 | 92.80 |
| 70 | 95.30 | 94.80 | 93.30 |

The optimal recognition performance of the MLP is observed at a point, on the curve of Fig. 7, where the number of neurons in its hidden layer is set to 54 on Set#1. Similarly, for Set#3 the optimal recognition performance is achieved where the number of neurons in its hidden layer is also 54. On Set#2, the optimal recognition performance is observed with 50 hidden layer neurons. However, the average recognition performance among the three sets is best with 54 neurons in the hidden layer. Thus the number of neurons in the hidden layer of the MLP is finally fixed up to 54. With this, the design process is completed producing an MLP (88-54-10) for recognition of handwritten numerals on the basis of the feature set explained before. The average Recognition performance of this MLP on the test sets, as observed with 54 hidden neurons, is 95% after 3-fold cross validation of results.